\documentclass[letterpaper, 10 pt, conference]{ieeeconf}  

\IEEEoverridecommandlockouts                              

\usepackage{graphicx}
\usepackage{float}
\usepackage{amsmath}
\usepackage{amsfonts}
\usepackage{booktabs}
\usepackage{makecell}
\usepackage{url}

\title{\LARGE \bf
CAST: Collision-Aware Assembly with Construction Robots using Simultaneous Trajectory Estimation and Planning
}

\author{Karthik Shaji$^{1}$, Chisung Kim$^{2}$, John D'Amato$^{2}$, Edvard Bruun$^{2}$ and Frank Dellaert$^{3}$
\thanks{*This material is based upon work supported by the National Science Foundation Graduate Research Fellowship Program under Grant No. DGE-2039655. Any opinions, findings, and conclusions or recommendations expressed in this material are those of the author(s) and do not necessarily reflect the views of the National Science Foundation.}
\thanks{$^{1}$Karthik Shaji is with the Daniel Guggenheim School of Aerospace Engineering,
        Georgia Institute of Technology, Atlanta 30332, USA
        {\tt\small kshaji3@gatech.edu}}%
\thanks{$^{2}$Chisung Kim, John D'Amato, and Edvard Bruun are with the School of Civil and Environmental Engineering,
        Georgia Institute of Technology, Atlanta 30332, USA, USA
        }%
\thanks{$^{3}$Frank Dellaert is with the Institute of Robotics and Intelligent Machines,
        Georgia Institute of Technology, Atlanta 30332, USA, USA
       }%
}

\begin{document}

\maketitle
\thispagestyle{empty}
\pagestyle{empty}

\begin{abstract}
Multi-robot systems have shown increasing viability in construction due to their ability to execute high-precision actions while reducing human exposure to hazardous tasks. However, these environments have high-dimensional configuration spaces and possess substantial collision-avoidance constraints, which include other robots, assembly objects, and workspace boundaries. We utilize a single factor graph for trajectory estimation and planning that incorporates measured robot states together with explicit collision and learned cable constraints. This supports changing workspaces and enables synchronized, high-dimensional robot motion planning while accounting for the stiff, vibration-induced uncertainty of heavy robotic systems. We demonstrate the success of our framework on the construction of a post-and-lintel structure using one robot arm as a timber gripper, and a second robot as a nail-fastener.
\end{abstract}

\section{Introduction}
Construction robots have shown promise to automate construction due to their high precision and ability to replace humans in dangerous tasks. Applications include block-block construction~\cite{Bruun21AIC_masonry}, additive manufacturing~\cite{Huang21CR_additive_bar}, and fastening tasks~\cite{Peters25ICRA_assembly_voxels}. Their ability to operate in constrained workspaces offers potential to streamline building of large-scale structures, for both terrestrial and planetary applications~\cite{Culbertson19IROS_space_assembly}. In recent years, interest in collaborative manufacturing has grown due to how it enables multiple actuators to support each other and work objects while carrying out a given set of tasks~\cite{Bruun22AIC_spaceframes}. For instance, robots together can lay bricks in a step-by-step method to build masonry arches~\cite{Bruun21AIC_masonry}, using methods of supporting each other, and thus avoiding putting humans into the construction workspace. 

However, to scale efficiently into real-world construction workspaces, these systems must also handle complex collision constraints, which includes robot-object, and robot-robot~\cite{Chen22Arxiv_coop_task} in the case of multi-actuator and manipulator systems. Existing approaches include Logic Geometric Programming (LGP)~\cite{Hartmann20IROS_long_horz_plan}, hierarchical methods with A*~\cite{Nagele20IROS_LegoBot}, and reinforcement learning~\cite{Cebula24IROS_feasible_assembly}. However, the LGP method~\cite{Hartmann20IROS_long_horz_plan} solves a joint nonlinear trajectory optimization problem over the entire task space, which can become computationally expensive as the planning horizon, number of discrete modes, and robot degrees of freedom (DoF) increase. Methods using A* are highly reliant on careful heuristic design, and can incur substantial runtime costs when backtracking is needed over large discrete search spaces. Lastly, the reinforcement learning method proposed in~\cite{Cebula24IROS_feasible_assembly} is itself not a continuous multi-robot motion planner, and rather relegates actual motion feasibility to inverse kinematics, collision checking, and cuRobo planning.
\begin{figure}
    \centering
    \includegraphics[width=0.8\linewidth]{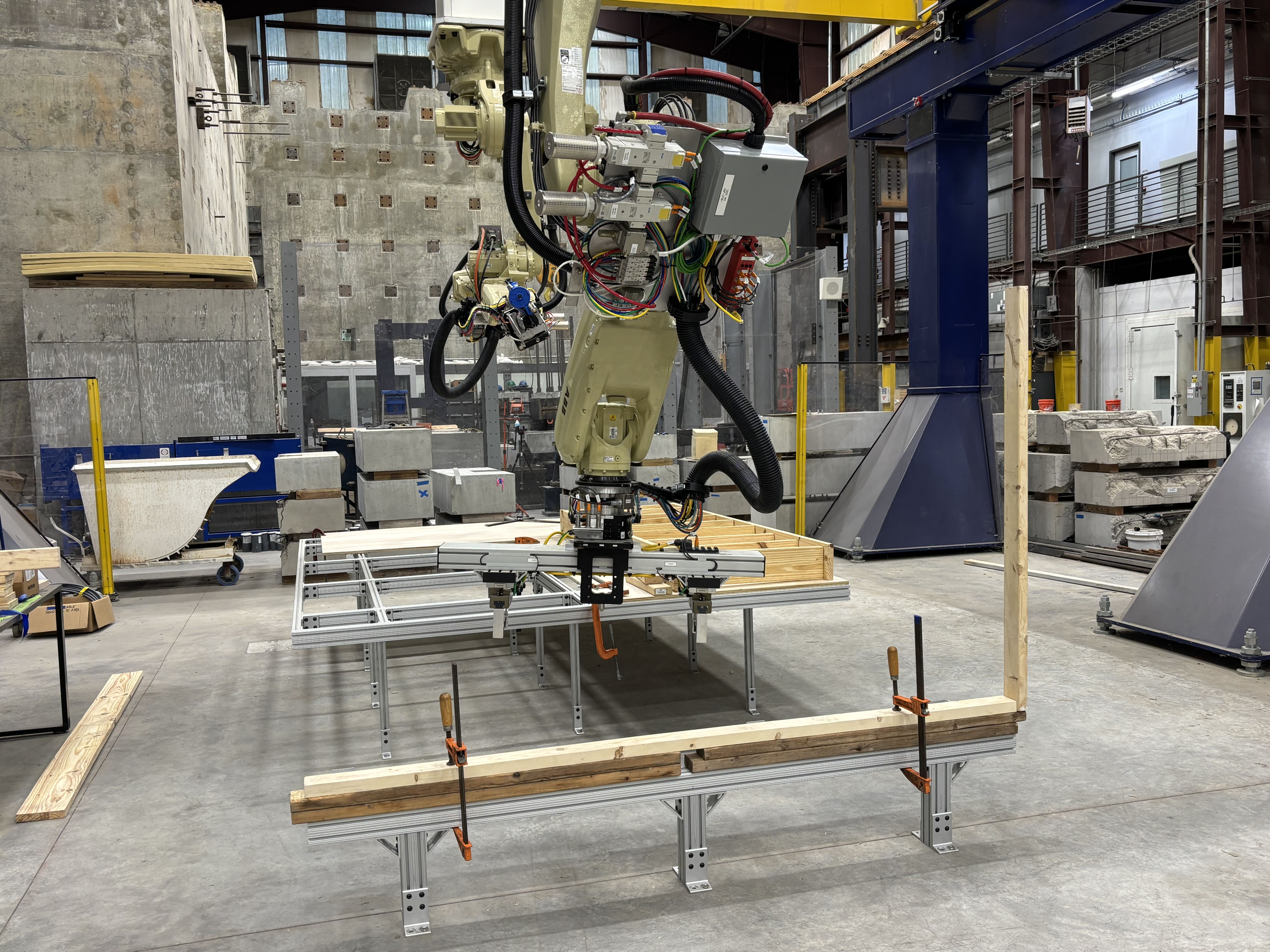}
        \caption{Dual-arm gantry and manipulator setup which holds a timber gripper on one arm and a nail gun on the other.}
    \label{fig:cmrg_irl}
\end{figure}
\indent Furthermore, these algorithms must efficiently handle the tradeoffs between roadmap preprocessing and single-query planning for collaborative assembly planning frameworks. While~\cite{Chen22Arxiv_coop_task} use multi-modal probabilistic roadmaps (PRMs)~\cite{Kavraki96TRO_prm} to enable efficient online task and path search, the required offline roadmap construction and collision annotation are computationally expensive, with collision annotation being found to require hundreds of hours for their larger domains. Recent approaches~\cite{Hartmann20IROS_long_horz_plan} have instead incorporated sampling-based single-query planners with Rapidly Exploring Random Tree (RRT) algorithms within hierarchical TAMP formulations to avoid constructing dense global roadmaps, which allow for rapid feasible path generation over complex configuration spaces. 

The ability for rapid, feasible path generation is critical in changing construction workspaces, where newly placed and assembled components change the collision environment. Approaches such as~\cite{Yang26RCM_decent_plan} used decentralized frameworks with usage of large language models (LLMs) for task planning. However, these methods introduce high-level task planning overhead and do not directly address challenges with high-dimensional collision optimization considered here. Many existing works~\cite{Chen22Arxiv_coop_task} treat the workspace as static and compute long-horizon trajectories based on assumed placement, but this may not be the case all the time. 

We treat our multi-actuator system as a single robotic system, building a unified factor graph that handles trajectory estimation, obstacle collision and self-collision constraints by stacking both robots into one state. We first compute an offline trajectory plan and then iteratively resolve it to estimate the current trajectory through the addition of measurements. We initialize the offline trajectory plan via RRT-Connect~\cite{Kuffner00ICRA_RRT_connect}, hence avoiding computational bottlenecks from extensive preprocessing and edge-validation steps in PRMs. We demonstrate the utility of our approach through simulation and hardware in the construction of a post-and-lintel structure. Furthermore, we show how our algorithm, CAST, can incorporate learned constraints obtained from deformable cable dynamics.

The key contributions of our work are listed below:
\begin{enumerate}
    \item We present a trajectory estimation and planning framework that uses a unified factor graph to incorporate measured robot states during execution and optimize future motion under collision and cable constraints.
    \item We build upon existing inter- and intra-robot collision formulations by introducing differentiable learned collision constraints that account for deformable cable behavior during motion.
    \item We demonstrate the proposed framework on an 18-DoF robotic system consisting of two actuators performing collaborative gripping and nail-fastening tasks to construct a post-and-lintel structure.
\end{enumerate}
An open source implementation of the code will be made available upon acceptance.

\section{Related Work}
\subsection{Construction Robotics/Cooperative Assembly}
Cooperative robotic fabrication combines component placement and temporary support within a shared workspace. Two-robot workflows have previously fabricated timber frame modules~\cite{Thoma18ROBARCH_timber} and spatial metal structures~\cite{Parascho18AAG_spatial}, showing capabilities across different materials and assembly geometries. In the domain of masonry construction, cooperative placement and support has enabled scaffold-free arch fabrication~\cite{Bruun21AIC_masonry}. Using multiple robots also supports reversible construction, such as in the ZeroWaste project which used a three-robot cell for the disassembly and reuse of timber framing~\cite{Bruun24CR_zerowaste}. Across these examples, robot motions must accommodate both the work of other robots and account for the partially assembled structure. However, while prior systems demonstrate cooperative fabrication, they primarily aim to show feasible techniques; we build on this by tackling automated continuous motion planning with changing assembly geometry.

\subsection{Trajectory Planning}
There are several methods for planning initial trajectories, including PRMs~\cite{Kavraki96TRO_prm} and RRT-based methods~\cite{Kuffner00ICRA_RRT_connect}. The advantage of PRM-based approaches is that they sample the configuration space to create a reusable roadmap, but this trades a higher initial runtime for a lower long-term runtime. When the collision geometry changes between assembly stages, this roadmap preprocessing and edge validation is harder to amortize. As a result, the benefits arising from PRMs are reduced as RRT methods can compute feasible paths more rapidly as they do not depend on this environment amortization, thus operating better in dynamic environments where multiple queries are not as necessary. However, methods such as RRT-Connect~\cite{Kuffner00ICRA_RRT_connect} do suffer from the drawback that they have no asymptotic optimality guarantee. Since this is not necessary for our approach as we are using it for setting a feasible seed, we choose it over PRMs.

\subsection{Factor Graph Motion Planning}
Factor graphs~\cite{Dellaert17_FactorGraphs,Dellaert12GTSAM_factorgraphs} have been used in motion planning for both Gaussian Process Motion Planning (GPMP)~\cite{Mukadam18IJRR_GPMP}, as well as for Simultaneous Trajectory Estimation and Planning (STEAP)~\cite{Mukadam18CoRR_steap}. The core advantage of this approach is that it leverages how Gaussian processes can be parameterized by a sparse set of support states and then be efficiently interpolated for collision checking in between nodes. When combined with factor graphs, the resultant sparse linear algebra becomes very efficient. Subsequent work on STEAP~\cite{KingSmith22ICRA_SCATE} has extended it to be capable of tracking moving objects. Our CAST framework follows the joint estimation-and-planning structure of STEAP while extending the factor graph to coordinated dual-robot collision optimization, changing assembly geometry, and learned cable constraints. Prior work on GPMP has combined sampling and trajectory optimization over graphs, allowing for improved efficiency in higher dimensions~\cite{Alwala21IROS_jsmp}. However, these works do not address the combination of joint inter- and intra-robot collision optimization, changing assembly geometry, and learned constraints.

\subsection{Deformation-Aware Planning}
Previous work in deformation-aware planning has addressed deformable mechanics either through component-level physical modeling or manipulation-based identification of material properties. Gayle et al.~\cite{Gayle05RSS_deformable} represented deformable systems through spring-mass-damper models. While these approaches provide a physics-level description, they require explicit mechanical characterization, and may be computationally costly. More recent work has considered deformation through manipulation, such as in~\cite{Aksoy26TASE_DLO}, where parameters such as Young's modulus were estimated using force and torque measurements. Yin et al.~\cite{Yin21SciRob_deformable} survey learning, perception, and control methods for deformable-object handling instead of through direct system property characterization. However, in our case the robot-mounted cable is not a manipulation or control target, and hence is not independently actuated or directly instrumentable.
\section{Preliminaries}

\subsection{Gaussian Processes for Trajectory Smoothing}
Vector-valued Gaussian processes (GPs) are a principled way of reasoning about continuous-time trajectories, where trajectories map time to state. We consider continuous-time trajectories as samples from a vector-valued GP,
\begin{equation}
\label{GP_equation}
    \boldsymbol{\theta}(t) \sim \mathcal{G}\mathcal{P}(\boldsymbol{\mu}(t), \mathcal{K}(t, t')),
\end{equation}
where $\boldsymbol{\mu}(t)$ is a vector-valued mean function and $\mathcal{K}(t, t')$ is a matrix-valued covariance function. The GP defines a prior over the space of trajectories, as in Eq.~\eqref{GP_prior}:
\begin{equation}
\label{GP_prior}
    p(\boldsymbol{\theta}) \propto \exp \biggl\{-\frac{1}{2} \lVert \boldsymbol{\theta} - \boldsymbol{\mu} \rVert_{\mathcal{K}}^2\biggr\}.
\end{equation}

A linear time-varying stochastic differential equation (LTV-SDE) can generate a structured kernel~\cite{Barfoot14RSS_BatchCT}, which we define as
\begin{equation}
    \dot{\boldsymbol{\theta}}(t) = A(t)\boldsymbol{\theta}(t) + \mathbf{u}(t) + F(t)\mathbf{w}(t),
\end{equation}
where $\mathbf{u}(t)$ is the known control input, $A(t)$ and $F(t)$ are time-varying matrices of the system, and $\mathbf{w}(t)$ is a white process-noise term modeled as the zero-mean GP
\begin{equation}\label{white_process}
    \mathbf{w}(t) \sim \mathcal{G}\mathcal{P}(\mathbf{0}, Q_c\delta(t - t')).
\end{equation}

As shown in~\cite{Mukadam18IJRR_GPMP, Mukadam18CoRR_steap, Barfoot14RSS_BatchCT}, Gaussian processes can be parameterized by sparse support states and queried at arbitrary times through GP interpolation. The posterior mean of the trajectory at time $\tau$ is obtained by conditioning on the support states,
\begin{equation}\label{GP_interp_eq}
    \boldsymbol{\theta}(\tau) = \bar{\boldsymbol{\mu}}(\tau) + \widetilde{\mathcal{K}}(\tau, t)\widetilde{\mathcal{K}}^{-1}(\boldsymbol{\theta} - \widetilde{\boldsymbol{\mu}}),
\end{equation}
which can be computed efficiently by exploiting the sparse GP-prior structure,
\begin{equation}
    \boldsymbol{\theta}(\tau) = \widetilde{\boldsymbol{\mu}}(\tau) + \Lambda(\tau)(\boldsymbol{\theta}_i - \widetilde{\boldsymbol{\mu}}_i) + \Psi(\tau)(\boldsymbol{\theta}_{i + 1} - \widetilde{\boldsymbol{\mu}}_{i + 1}).
\end{equation}

\subsection{Gaussian Processes in Factor Graphs}
A factor graph is a bipartite graph $G = \{\Theta, \mathcal{F}, \mathcal{E}\}$, where $\Theta$ denotes variable nodes, $\mathcal{F}$ denotes factor nodes, and $\mathcal{E}$ denotes the edges connecting them. In GPMP, the variable nodes correspond to trajectory states, while each factor encodes a probabilistic cost or constraint. The posterior distribution can thus be written as
\begin{equation}
    p(\boldsymbol{\theta} \mid \mathbf{e}) \propto \prod_{m=1}^{M} f_m(\Theta_m),
\end{equation}
where $f_m$ are factor terms on variable subsets $\Theta_m$. Within GPMP~\cite{Mukadam18IJRR_GPMP}, obstacle and joint-limit constraints are standard factor terms used to encode collision avoidance and feasible joint ranges. The obstacle factor is defined as
\begin{equation}
\label{obstacle_sdf}
    f_i^{\mathrm{obs}}(\boldsymbol{\theta}_i) = \exp\left\{-\frac{1}{2}\left\|\mathbf{h}(\boldsymbol{\theta}_i)\right\|_{\sigma_{\mathrm{obs}}}^2\right\}.
\end{equation}
We utilize a trilinearly interpolated signed distance field (SDF) so that the distance to the nearest obstacle and its gradient are available in closed form at any point. For each query point on the robot, determined by the current configuration, we evaluate a hinge loss on its distance $d$ to the relevant collision boundary,
\begin{equation}
\label{hinge_loss}
    c(d) =
    \begin{cases}
        d_{\mathrm{safe}} - d & \text{if } d < d_{\mathrm{safe}} \\
        0 & \text{otherwise},
    \end{cases}
\end{equation}
where $d_{\mathrm{safe}}$ is the standoff distance obtained by extending a safety margin $\epsilon$ (in practice, we use spheres as our collision primitive, so $d_{\mathrm{safe}} = r + \epsilon$). For obstacle avoidance, $d$ is the SDF distance at the query point; for self-collision, $d = \lVert \mathbf{x}_a - \mathbf{x}_b \rVert$ for a non-adjacent link pair $(a,b) \in \mathcal{S}$, with $d_{\mathrm{safe}} = r_a + r_b + \epsilon_{ab}$. Once a point passes this margin, the graph error grows quadratically with penetration depth, driving the optimizer back onto the safe path.

Joint-limit violations are handled by projecting violating states back into the feasible region. These factors are used in the resulting optimization problem in our work. A GP-interpolated obstacle factor also allows collision checking between support states,
\begin{equation}
\begin{aligned}
    f_{\tau_j}^{\mathrm{intp}}(\boldsymbol{\theta}_i, \boldsymbol{\theta}_{i+1})
    &= \exp\left\{-\frac{1}{2}\left\|\mathbf{h}\bigl(\boldsymbol{\theta}(\tau_j)\bigr)\right\|_{\sigma_{\mathrm{obs}}}^2\right\} \\
    &= \exp\left\{-\frac{1}{2}\left\|\mathbf{h}_{\tau_j}^{\mathrm{intp}}(\boldsymbol{\theta}_i, \boldsymbol{\theta}_{i+1})\right\|_{\sigma_{\mathrm{obs}}}^2\right\}.
\end{aligned}
\label{eq:intp_factor}
\end{equation}
\section{Joint Estimation and Planning}
We formulate trajectory estimation and planning jointly as a single maximum a posteriori (MAP) inference over a factor graph. The MAP trajectory $\boldsymbol{\theta}^{*}$ maximizes the product of all factors, equivalently minimizing the sum of their weighted squared residuals:
\begin{equation}
    \boldsymbol{\theta}^{*}
    = \arg\max_{\boldsymbol{\theta}} \prod_{m} f_m(\boldsymbol{\theta})
    = \arg\min_{\boldsymbol{\theta}} \sum_{m} \left\lVert \mathbf{e}_m(\boldsymbol{\theta}) \right\rVert_{\boldsymbol{\Sigma}_m}^{2},
\label{eq:map}
\end{equation}
where each factor $f_m$ contributes a residual $\mathbf{e}_m$ weighted by its covariance $\boldsymbol{\Sigma}_m$. Measurement factors perform estimation, while GP-prior, goal, obstacle, self-collision, and cable factors perform planning; feasibility requirements such as joint limits, obstacle clearance, and fixed start and goal states enter as factors in this objective rather than as hard constraints.

\subsection{CAST Factor Graph}
We model the dual-arm system with an 18-dimensional state vector including both actuators and gantries, allowing for optimization across the full system with a single factor graph. The stacked configuration is $\boldsymbol{\theta} = [\boldsymbol{\theta}^{1\top}, \boldsymbol{\theta}^{2\top}]^{\top} \in \mathbb{R}^{D}$, with $D = 18$, ensuring that all constraints are optimized jointly rather than per-arm. Robot-robot collision is handled within the same stacked state representation, since both actuators are included in a single configuration vector. We use the self-collision factor to handle link-link collision checking across both arms, which allows for a single MAP inference over the graph coordinates to coordinate both arms. In practice, our robot collision geometry is represented with spheres, although the formulation below is stated more generally.

In the factor graph implementation, the obstacle, self-collision, and cable factors are finite-weight soft penalties in the MAP objective. They encourage clearance, but do not provide hard constraints, and the resulting MAP estimate includes an estimate of the executed trajectory and a plan for the remaining motion.

\subsubsection{Measurement Factors}
Following the measurement-factor structure used in STEAP~\cite{Mukadam18CoRR_steap}, we incorporate measured robot states into the factor graph during maneuver execution. For a measured state $\mathbf{z}_i$ associated with trajectory state $\boldsymbol{\theta}_i$, we define the measurement factor as
\begin{equation}
    f_i^{\mathrm{meas}}(\boldsymbol{\theta}_i)
    = \exp\left\{
    -\frac{1}{2}
    \left\|
    \mathbf{r}_i(\boldsymbol{\theta}_i, \mathbf{z}_i)
    \right\|_{\boldsymbol{\Sigma}_i}^{2}
    \right\},
\label{eq:meas_factor}
\end{equation}
where $\mathbf{r}_i$ is the measurement residual and $\boldsymbol{\Sigma}_i$ is the measurement covariance. These factors update the posterior trajectory estimate as measurements become available.

\subsubsection{Self-Collision Factors}
Self-collision is penalized through a hinge-loss factor over non-adjacent link pairs, thereby optimizing for link-link clearance during motion, as shown in Eq.~\eqref{eq:self_intp_factor}. Following~\cite{Mukadam18IJRR_GPMP, Alwala21IROS_jsmp}, we represent a robot body with query points $\phi(\boldsymbol{\theta}) = \{\mathbf{x}_1, \ldots, \mathbf{x}_P\}$ computed from the joint-space configuration via forward kinematics. Self-collision is only possible for non-adjacent links; therefore, we only evaluate constraints for link pairs with non-constant separation, which we do via the hinge loss in Eq.~\eqref{hinge_loss}. An SDF is not required here because the distance is evaluated between body points that directly depend on the configuration. As done in~\cite{Alwala21IROS_jsmp}, we stack the losses of all pairs into $\mathbf{h}^{self}(\boldsymbol{\theta}_i)$, yielding the unary self-collision factor
\begin{equation}
    f_i^{self}(\boldsymbol{\theta}_i) =
    \exp\left\{-\frac{1}{2}\left\|\mathbf{h}^{self}(\boldsymbol{\theta}_i)\right\|_{\sigma_{self}}^2\right\}.
\end{equation}
Like its unary obstacle counterpart, this only penalizes self-collision at the nodes, and neglects the trajectory between them.

Therefore, we introduce a new GP-interpolated factor for self-collision, constructed analogously to Eq.~\eqref{eq:intp_factor}, where the state $\boldsymbol{\theta}(\tau_j)$ is recovered from the neighboring support states, and the self-collision loss is evaluated there,
\begin{equation}
\begin{aligned}
    f_{\tau_j}^{self,intp}(\boldsymbol{\theta}_i,\boldsymbol{\theta}_{i+1})
    &=\exp\left\{-\frac{1}{2}\left\|\mathbf{h}^{self}\bigl(\boldsymbol{\theta}(\tau_j)\bigr)\right\|_{\sigma_{self}}^2\right\} \\
    &=\exp\left\{-\frac{1}{2}\left\|\mathbf{h}_{\tau_j}^{self,intp}(\boldsymbol{\theta}_i,\boldsymbol{\theta}_{i+1})\right\|_{\sigma_{self}}^2\right\}.
\end{aligned}
\label{eq:self_intp_factor}
\end{equation}
This binary factor connects consecutive support states and penalizes self-collision between them, which improves continuous-time clearance for articulated robots.
\subsection{Initialization}
RRT-Connect~\cite{Kuffner00ICRA_RRT_connect} provides the initial feasible trajectory used to initialize the factor graph, after which we refine it through joint optimization over smoothness, self-collision, and cable-spline constraints. Gaussian-process motion planning is non-convex, so the initialized trajectory strongly affects the resulting optimum. RRT-Connect is a single-query planner that grows two trees iteratively: one rooted at the start and one rooted at the goal, and greedily connects them. To obtain the start and goal states, we use inverse kinematics on the world-space coordinates. The resulting waypoint path is then resampled to the $n$ support states used in GPMP and used as the initial estimate. As GPMP optimizes both smoothness and clearance, we only require an approximate trajectory; in particular, we initialize the trajectory to be feasible for the SDF obstacle field while leaving the cable-collision and self-collision to be handled by GPMP.

At each assembly step, the workspace changes as components are placed; accordingly, a new initialization is performed using the updated obstacle set, and the resulting factor graph is solved. We also considered RRT$^x$~\cite{Otte116IJRR_RRTX}, which is designed to repair the graph in the presence of continuously moving obstacles. However, while that approach may be appropriate for GPMP methods such as~\cite{KingSmith22ICRA_SCATE}, the environment here changes only between discrete assembly steps, and each step is planned against a fixed obstacle snapshot. This makes the asymptotically optimal machinery in that approach less useful in our setting, motivating the use of RRT-Connect.

\subsection{Online Estimation and Replanning}
During execution, measured robot states, which are joint angles and gantry positions, are incorporated through measurement factors, updating the posterior estimate of the trajectory. The updated estimate is then used to generate the remaining motion plan, in STEAP fashion~\cite{Mukadam18CoRR_steap}.
\section{Learned Cable-Spline Factors}
A key feature of our system is how we model deformable cables attached to the actuators with a learned spline representation and integrate it into the same factor graph optimization framework. A significant challenge in the construction-robotics setting is the presence of a large, deformable cable housing that moves with the actuators as the joint configuration changes. If this cable housing snags on obstacles or on another actuator, it can cause substantial damage to the robot and prevent the task from being completed. Traditional cable models are often based on finite-element or related physics-based solvers, but these methods are computationally expensive. We therefore learn a neural-network model of the cable geometry to enable fast inference during optimization.
\subsection{Modeling Cable Geometry as a Chebyshev Spline}
To model the cable geometry itself, we use a pseudospectral Chebyshev spline. Rather than parameterizing a polynomial by its coefficients, the pseudospectral approach parameterizes it by its values at a set of Chebyshev points of the second kind. Mapped onto the unit interval $s \in [0,1]$, these points are given by
\begin{equation}
    s_j = \frac{1}{2}\left(1 - \cos\frac{j\pi}{N-1}\right), \qquad j = 0,\ldots,N-1,
\label{eq:cheb_points}
\end{equation}
which cluster near the endpoints of the interval, avoiding the oscillations that arise when interpolating at equally spaced nodes. Given nodal values $\mathbf{p}_j = \mathbf{p}(s_j)$, the spline can then be evaluated at any $s \in [0,1]$ via barycentric Lagrange interpolation \cite{Berrut04_Lagrange}:
\begin{equation}
    \mathbf{p}(s) = \sum_{j=0}^{N-1} w_j(s)\,\mathbf{p}_j,
    \quad
    w_j(s) = \frac{\lambda_j / (s - s_j)}{\sum_{k=0}^{N-1} \lambda_k / (s - s_k)},
\label{eq:barycentric}
\end{equation}
where $\lambda_j = (-1)^j$, halved at the two endpoint nodes $j = 0$ and $j = N-1$. The weights satisfy $w_j(s_k) = 1$ if $j = k$ and $0$ otherwise, so the interpolant passes exactly through its nodal values. As the evaluation is linear in the nodal values, the derivatives required for gradient-based trajectory optimization are readily available.

We use this representation as the output parameterization of the neural network. The two cable attachment points are assumed to be known because each is fixed in the local frame of its corresponding robot link and can therefore be transformed into the selected reference frame. For a configuration $\boldsymbol{\theta}$, let $\mathbf{p}_A(\boldsymbol{\theta})$ and $\mathbf{p}_B(\boldsymbol{\theta})$ denote the attachment points anchored at the endpoint nodes $s_0 = 0$ and $s_{N-1} = 1$. Let $\mathbf{R}(\boldsymbol{\theta})$ denote the rotation from the prediction frame to the selected reference frame, and let $\boldsymbol{\rho}(\boldsymbol{\theta}) = g_{\boldsymbol{\phi}}(\mathbf{S}(\boldsymbol{\theta}))$ denote the learned multilayer perceptron with parameters $\boldsymbol{\phi}$. The selection operator $\mathbf{S}$ identifies the joint variables supplied to the network. The network predicts the $N-2$ interior Chebyshev nodal residuals $\boldsymbol{\rho} = [\boldsymbol{\rho}_1, \ldots, \boldsymbol{\rho}_{N-2}]$, where $\boldsymbol{\rho}_j \in \mathbb{R}^3$, relative to the chord between the attachment points. The predicted cable position $\mathbf{x}^{cab}$ is
\begin{equation}
    \begin{aligned}
        \mathbf{x}^{cab}(s;\boldsymbol{\theta}) ={} & (1-s)\,\mathbf{p}_A(\boldsymbol{\theta}) + s\,\mathbf{p}_B(\boldsymbol{\theta}) \\
        & + \mathbf{R}(\boldsymbol{\theta})\sum_{j=1}^{N-2} w_j(s)\,\boldsymbol{\rho}_j(\boldsymbol{\theta}),
        \qquad s \in [0,1],
    \end{aligned}
\label{eq:cable_curve}
\end{equation}
where $w_j(s)$ are the barycentric weights of Eq.~\eqref{eq:barycentric}. Since the residual vanishes at the attachment points, the sum runs over the interior nodes only.

\subsection{Cable Spline Factor}
We incorporate the learned cable geometry into a cable-spline factor. The factor evaluates the cable curve in Eq.~\eqref{eq:cable_curve} at a given configuration $\boldsymbol{\theta}$. The cable cost vector stacks the sphere-based hinge loss $\mathbf{h}(\cdot)$ with standoff inflated by the cable radius $r_{cab}$ at $M$ samples $\{s_m\}$ along the curve:
\begin{equation}
    \mathbf{h}^{cab}(\boldsymbol{\theta}) = \bigl[\, \mathbf{h}\bigl(\mathbf{x}^{cab}(s_1;\boldsymbol{\theta})\bigr),
    \; \ldots, \; \mathbf{h}\bigl(\mathbf{x}^{cab}(s_M;\boldsymbol{\theta})\bigr) \,\bigr]^{\top}.
\end{equation}
The unary cable-spline factor follows the same form as the standard obstacle factor:
\begin{equation}
    f_i^{cab}(\boldsymbol{\theta}_i) = \exp\left\{-\frac{1}{2}\left\|\mathbf{h}^{cab}(\boldsymbol{\theta}_i)\right\|_{\sigma_{cab}}^2\right\}.
\label{eq:cable_factor}
\end{equation}
The corresponding Gaussian-process interpolated cable factor is
\begin{equation}
\begin{aligned}
    f_{\tau_j}^{cab,intp}(\boldsymbol{\theta}_i,\boldsymbol{\theta}_{i+1})
    &= \exp\left\{-\frac{1}{2}\left\|\mathbf{h}^{cab}\bigl(\boldsymbol{\theta}(\tau_j)\bigr)\right\|_{\sigma_{cab}}^2\right\} \\
    &= \exp\left\{-\frac{1}{2}\left\|\mathbf{h}_{\tau_j}^{cab,intp}\right\|_{\sigma_{cab}}^2\right\}.
\end{aligned}
\label{eq:cable_intp_factor}
\end{equation}
\section{Experimental Setup}
\label{sec:experimental_setup}
\subsection{Robot Description}
\begin{figure}
        \centering
        \includegraphics[width=0.9\columnwidth]{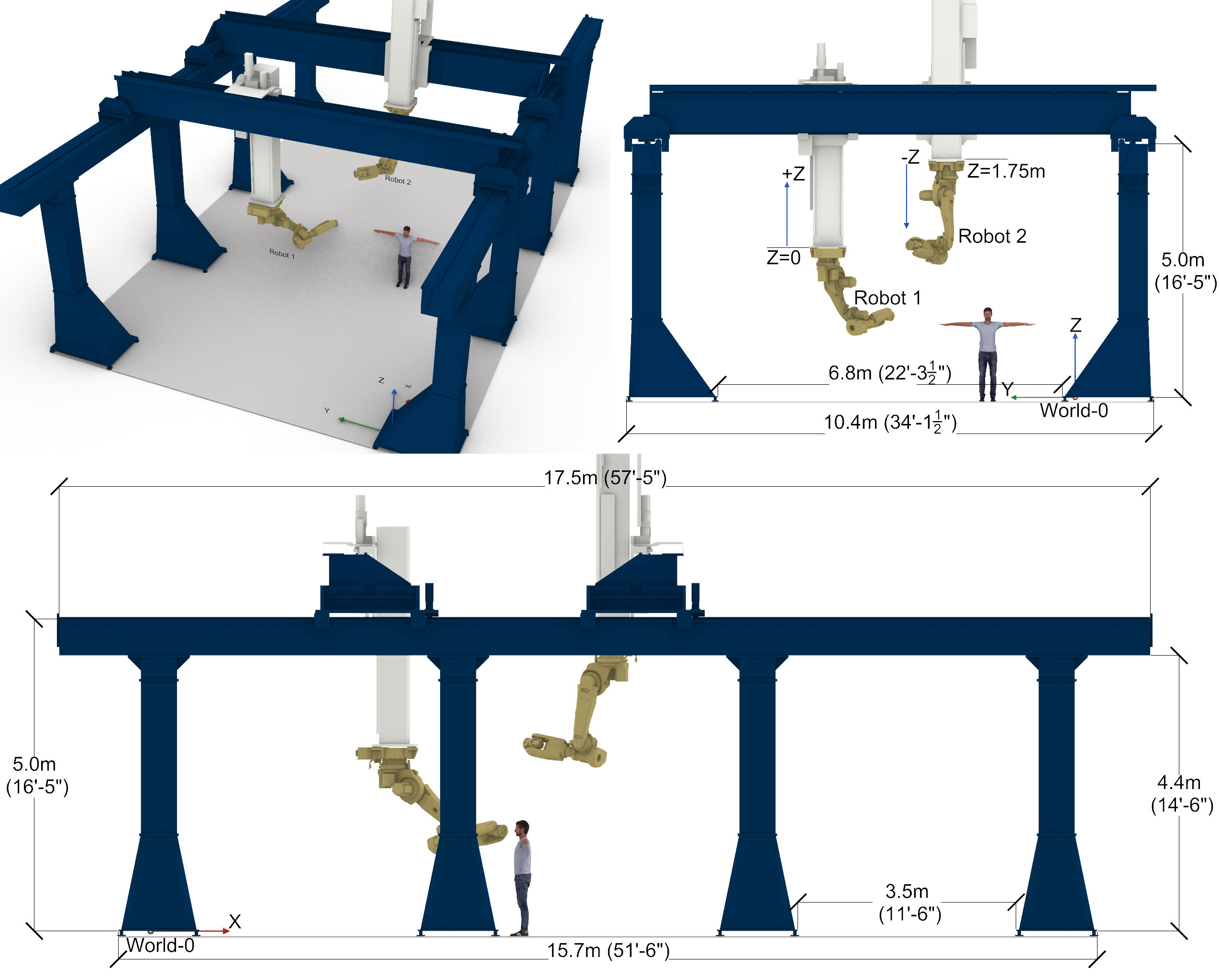}
        \caption{System arrangement and dimensions of our 18-DoF construction setup.}
        \label{fig:18-DoF}
\end{figure}

We use a dual-arm construction system consisting of two 6-DoF manipulators, whose arrangement and dimensions are shown in Figs.~\ref{fig:cmrg_irl} and~\ref{fig:18-DoF}. We use COMPAS Fab~\cite{compas-fab} for the control software on the robot and COMPAS RRC~\cite{compas-rrc} to interface with it in Python. Each arm is mounted on a 3-DoF gantry, yielding a 9-DoF arm configuration that provides large-scale positioning in the x-, y-, and z-directions while preserving fine-grained end-effector control. Each arm can be equipped with either a nail gun or a gripper for handling timber. The experiments are designed to demonstrate coordinated multi-actuator motion planning with trajectory estimation in a dynamically changing construction workspace. Two robots cooperatively handle and fasten components during assembly. To evaluate the planner and estimator, each assembly step begins without a complete model of the structure that will result from the preceding steps.

\subsection{Simulation and Algorithm Implementation}
Before testing our code on real hardware, we test our manufacturing sequences in both Rhino/Grasshopper/Kangaroo, and RobotStudio for manual collision checking prior to deployment.  We trained our neural network for cable dynamics on a Kangaroo Extended Position-Based Dynamics (XPBD) solver~\cite{Macklin16ICMG_xpbd}, and ran simulations on 4096 rested cable poses on different joint configurations for each arm with their respective tool attached. The algorithm development was done in Python using the open source libraries GTSAM~\cite{gtsam} and GTDynamics with Python bindings. 

For integrating measurement factors into motion planning, we fold in new measurements of the 18-DoF robot state from per-joint sensors using iSAM2~\cite{Kaess12IJRR_isam2}. Each new factor is linearized at the current estimate, and iSAM2 then moves up from the affected variable's clique to the root, marking all cliques along that path. It then undoes those marked cliques back into a factor graph, re-eliminates it, and then back-substitutes to update affected variables. In this way, only affected cliques and trajectory values are changed.
\subsection{Experiments}
\begin{figure}[t]
        \centering
        \includegraphics[width=0.6\columnwidth]{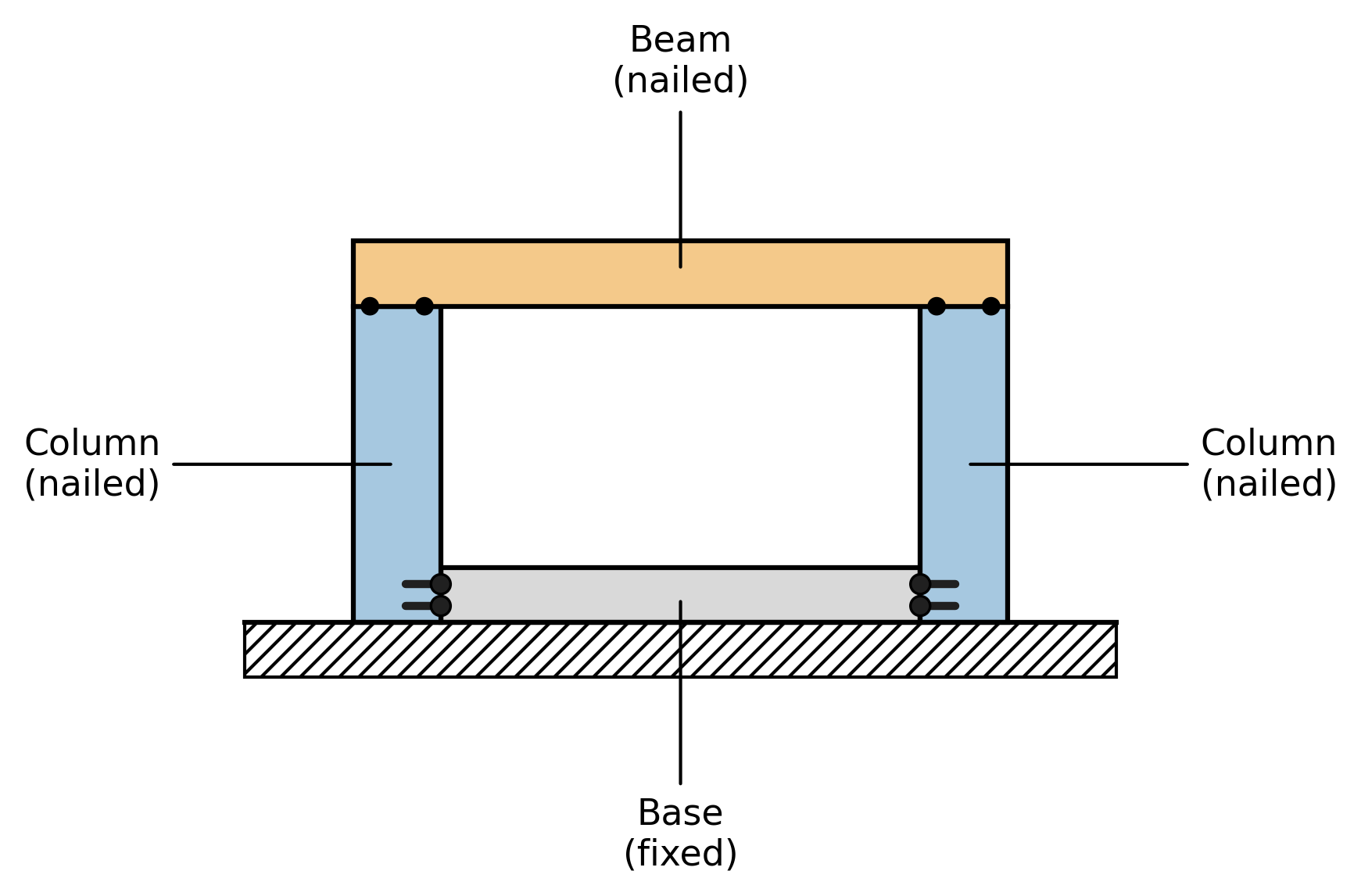}
        \caption{Post-and-lintel structure assembly, with annotations for fixed and nailed components.}
        \label{fig:lintel_structure}
\end{figure}

\begin{figure}[t]
        \centering
        \includegraphics[width=0.85\columnwidth]{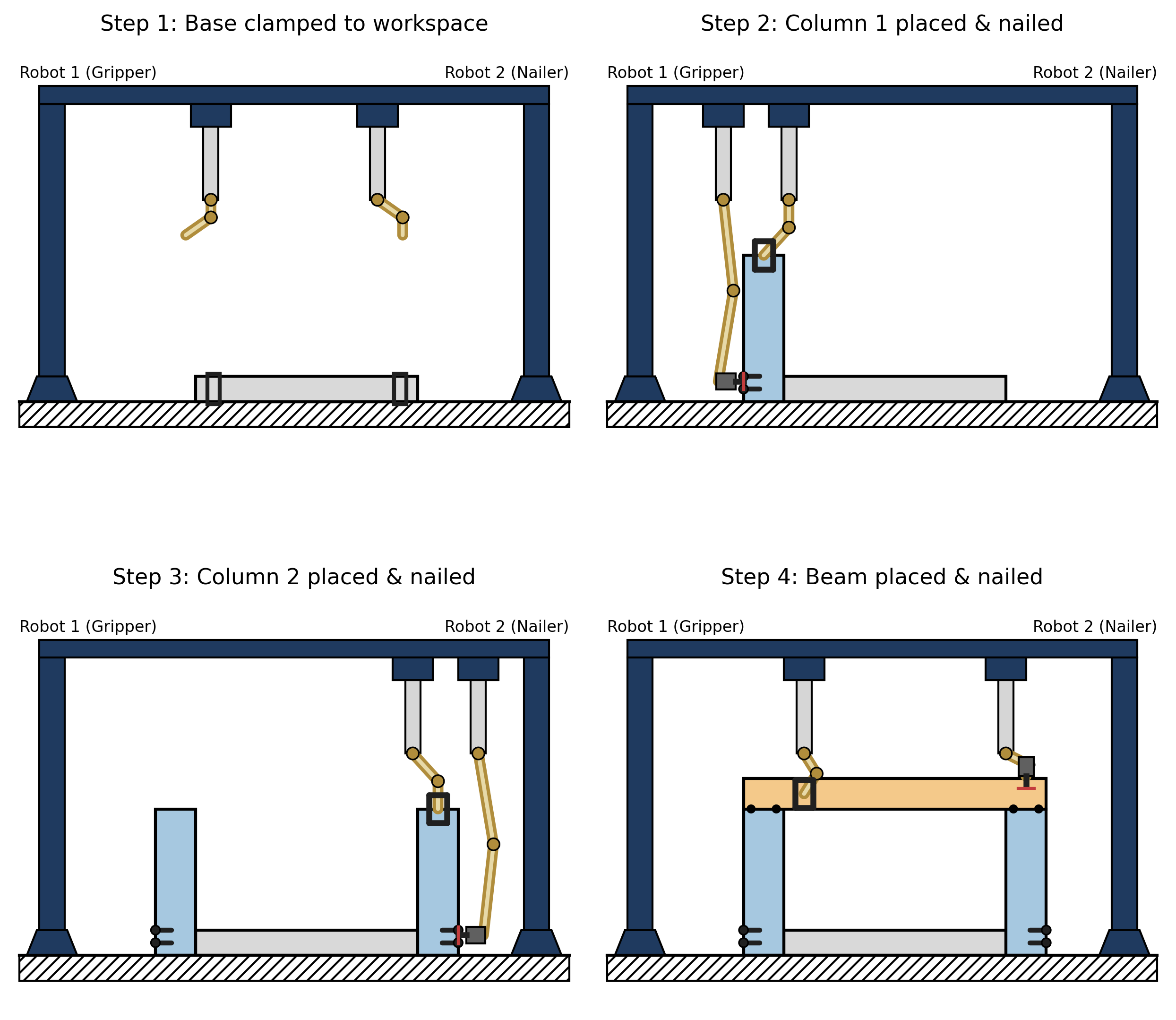}
        \caption{Assembly sequence for the post-and-lintel structure. The base is first fixed to the workspace via clamps. Then robot 1 first grips the left column and then the right column while robot 2 performs the nailing. Finally, robot 1 places the final beam on top and robot 2 fastens it into place.}
        \label{fig:lintel_build_sequence}
\end{figure}

We use the assembly of a post-and-lintel structure as a representative test case, shown in Fig.~\ref{fig:lintel_structure}. This structure consists of a fixed base, two columns which are nailed to the base from the sides, and a top beam nailed across the column tops. After a component has been placed and fastened, the estimator incorporates measurements of the executed motion and the planner updates the workspace and collision constraints before generating the next motion plan. The new columns added to the workspace have known measurements and we assume they are placed where intended. This procedure exposes the system to changing obstacle and self-collision constraints while requiring the robots to coordinate their roles: one robot can support a component while the other performs the fastening operation. The resulting plans demonstrate simultaneous multi-actuator coordination in a progressively changing construction environment.
Fig.~\ref{fig:lintel_build_sequence} illustrates the four-step build order used to evaluate the planner and estimator. 

\section{Experimental Results}
\subsection{Simulation - Cable Dynamics}
We train one cable-geometry network per arm due to how each arm's tool makes the cable drape differently, and trade prediction accuracy against inference runtime. We present the results of our ablations on both arms in Tables~\ref{tab:ablation-arm1-gripper} and~\ref{tab:ablation-arm2-nailer}. We chose the bolded architecture, which had the most runtime, but highest accuracy.

\begin{table}[htbp]
\centering
\caption{Cable model ablations for Arm 1 (Gripper). Architecture is listed as (hidden layers) $\times$ (neurons per layer).}
\label{tab:ablation-arm1-gripper}
\footnotesize
\setlength{\tabcolsep}{4pt}
\begin{tabular}{@{}crrr@{}}
\toprule
Architecture & Train RMSE (cm) & Test RMSE (cm) & $\mu$s/sample \\
\midrule
$1\times16$ & 11.58 & 11.50 & 0.461 \\
$1\times32$ & 10.73 & 10.88 & 0.498 \\
$1\times64$ & 9.20  & 9.85  & 0.702 \\
$2\times16$ & 10.16 & 10.58 & 0.871 \\
$2\times32$ & 8.35  & 9.37  & 1.012 \\
$2\times64$ & 6.82  & 8.56  & 1.108 \\
$3\times16$ & 9.23  & 10.39 & 1.407 \\
$3\times32$ & 6.75  & 8.75  & 2.157 \\
$\mathbf{3\times64}$ & \textbf{5.21} & \textbf{7.81} & \textbf{2.242} \\
\bottomrule
\end{tabular}
\end{table}
\begin{table}[htbp]
\centering
\caption{Cable model ablations for Arm 2 (Nailer). Architecture is listed as (hidden layers) $\times$ (neurons per layer).}
\label{tab:ablation-arm2-nailer}
\footnotesize
\setlength{\tabcolsep}{4pt}
\begin{tabular}{@{}crrr@{}}
\toprule
Architecture & Train RMSE (cm) & Test RMSE (cm) & $\mu$s/sample \\
\midrule
$1\times16$ & 11.61 & 11.56 & 0.639 \\
$1\times32$ & 10.40 & 10.57 & 0.576 \\
$1\times64$ & 9.06  & 9.46  & 0.559 \\
$2\times16$ & 10.73 & 10.77 & 0.766 \\
$2\times32$ & 7.80  & 8.95  & 0.780 \\
$2\times64$ & 6.76  & 8.65  & 0.913 \\
$3\times16$ & 7.97  & 8.97  & 0.834 \\
$3\times32$ & 6.18  & 7.87  & 0.956 \\
$\mathbf{3\times64}$ & \textbf{4.59} & \textbf{7.60} & \textbf{1.516} \\
\bottomrule
\end{tabular}
\end{table}

\subsection{Planning and Estimation}
Trajectory generation using our framework is efficient across all three sub-assembly tasks, with RRT-Connect taking an average of 46.9 ms for initialization, GPMP generation taking an average of 46.7 s, and each iSAM2 update taking an average of 292 ms (Table~\ref{tab:steap-timing}).
\begin{table}[t]
\centering
\caption{Timing statistics for trajectory generation and estimation}
\label{tab:steap-timing}
\footnotesize
\setlength{\tabcolsep}{4pt}
\begin{tabular}{@{}lrrr@{}}
\toprule
Member & RRT Init (ms) & Offline Traj. Gen. (s) & iSAM2 Update (ms) \\
\midrule
Left Col. & 40.8 & 38.9 & 225 \\
Right Col. & 64.4 & 46.2 & 255 \\
Top Beam & 35.6 & 54.9 & 396 \\
\midrule
\textbf{Average} & 46.9 & 46.7 & 292 \\
\bottomrule
\end{tabular}
\end{table}
Initialization via RRT-Connect is a negligible component of the total runtime cost for the offline trajectory generation, whose reported statistics include the RRT-initialization time. The final approach and withdraw at each nail are short, single-axis gantry moves into an already-cleared target, so we script them; the collision-relevant transit and repositioning motion is planned. The beam repositions cross the structure, so they are planned with GPMP/iSAM2, which accounts for its higher computation time. The iSAM2 update time is two orders of magnitude lower than trajectory generation, and when empirically compared to trajectory execution time, our system experienced no noticeable lag in decision-time during real-time execution.

In Table~\ref{tab:steap-clearance}, we show how our algorithm maintains safe average minimum clearances for obstacle, cable, and self-collision constraints at every assembly stage, excluding the steps where the nail gun approaches to nail. While the column obstacle clearances are below our 10 cm clearance target, this is reflective of the soft penalty methods we used, and can also be attributed to the empirical conservativeness of our sphere collision model when comparing sphere representation vs. actual meshes.
\begin{table}[t]
\centering
\caption{Average minimum clearance per maneuver, for obstacle, cable, and self-collision constraints.}
\label{tab:steap-clearance}
\footnotesize
\setlength{\tabcolsep}{4pt}
\begin{tabular}{@{}lrrr@{}}
\toprule
Member & Obstacle (m) & Cable (m) & Self-Collision (m) \\
\midrule
Left Col. & 0.037 & 0.175 & 0.877 \\
Right Col. & 0.024 & 0.200 & 0.568 \\
Top Beam & 0.021 & 0.251 & 0.470 \\
\midrule
\textbf{Average} & 0.027 & 0.208 & 0.638 \\
\bottomrule
\end{tabular}
\end{table}
\subsection{Hardware Testing}
\begin{figure}
    \centering
    \includegraphics[width=0.9\linewidth]{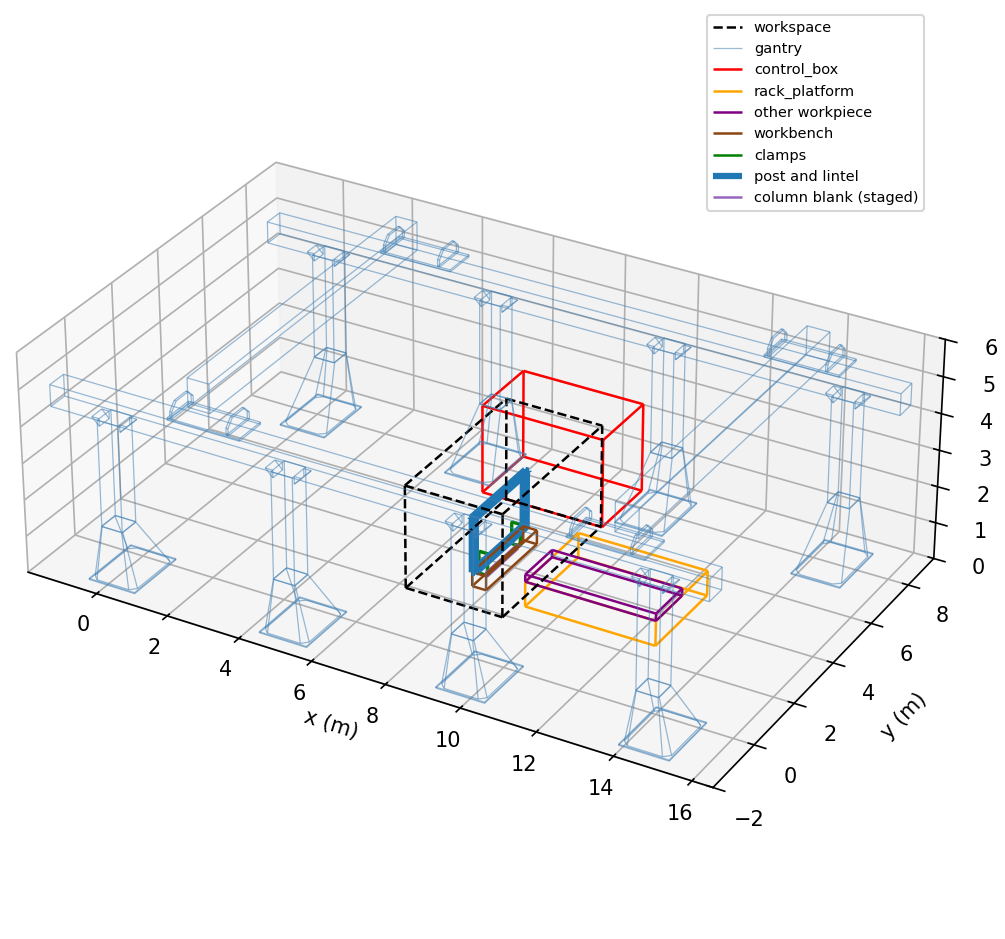}
    \caption{Tightly constrained workspace of hardware test including obstacles.}
    \label{fig:obstacle_env}
\end{figure}

\begin{figure}
    \centering
    \includegraphics[width=0.8\linewidth]{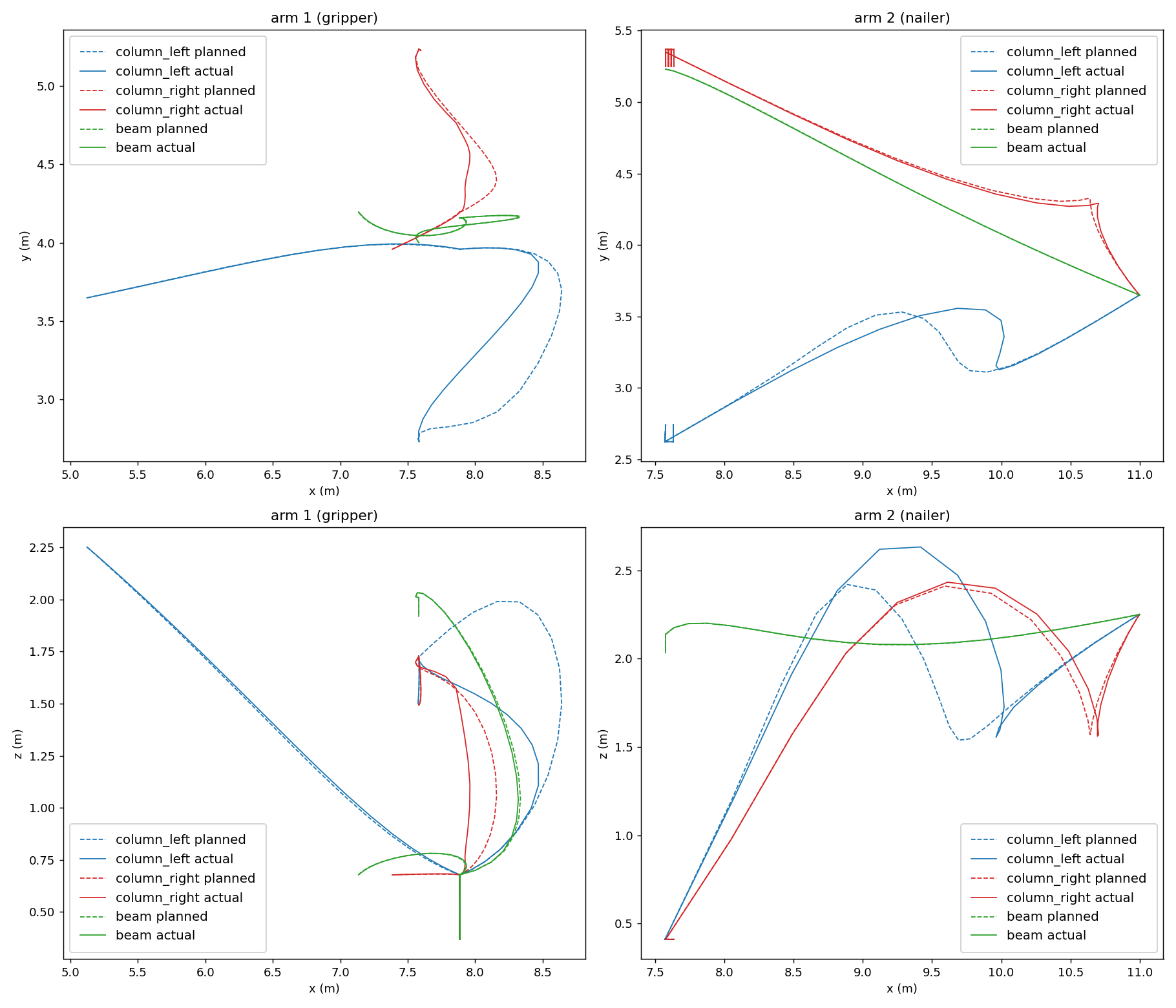}
    \caption{Preplanned (dashed) vs. executed (solid) end-effector centroid trajectories per arm, shown as $xy$, $xz$ coordinates.}
    \label{fig:end_effector_trajectories}
\end{figure}

\begin{figure}
    \centering
    \includegraphics[width=0.8\linewidth]{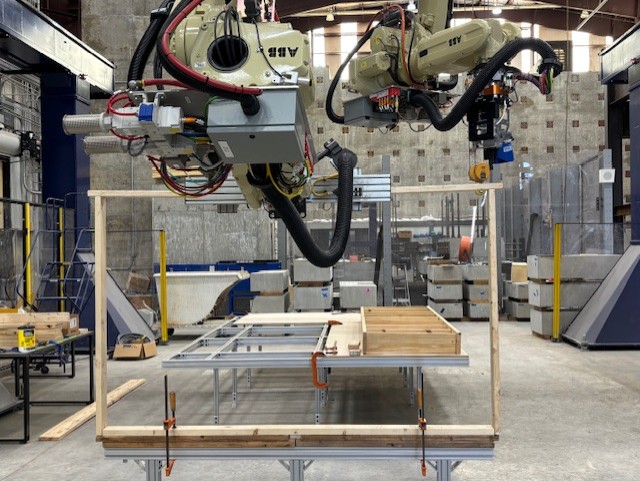}
    \caption{Completed post-and-lintel structure with both robot arms in their final configurations.}
    \label{fig:finished_object}
\end{figure}

We validate the full pipeline on hardware in a tightly constrained workspace, shown in Fig.~\ref{fig:obstacle_env}, planning each trajectory offline and updating it from live measurements through iSAM2 during execution.
For safety when working with large robotic setups, we broke our total trajectory into three, separately testable parts, and stored a trajectory checkpoint for reproducibility in the next run. Hence, once one assembly step was complete, we would test the next trajectory first, to account for effects not modeled in our simulation such as wood warping, bending, and shifting during placement, and then run it. We stitched the resulting trajectories together, along with the planned reference with respect to the end-effector centroid in Fig.~\ref{fig:end_effector_trajectories}. The executed trajectory closely follows the nominal plan, with divergences at looping repositioning segments; this is consistent with regions of higher uncertainty where iSAM2 triggers replanning. After nailing the first column, which only had two nails, we manually nailed in two extra nails for stability. We then updated the second column trajectory to nail four nails automatically. We show the finished product in Fig.~\ref{fig:finished_object}.

\section{Conclusion}
We presented CAST, a unified factor graph framework that jointly plans and estimates collision-free trajectories for a dual-arm construction system, combining explicit obstacle and self-collision factors with a learned cable-deformation constraint and live measurement updates. We implemented and evaluated our work on a task involving the building of a post-and-lintel structure in a highly constrained workspace. In future work, we aim to add sensing through cameras or LiDAR to actively sense and identify unexpected obstacles and changes in the workspace.



\section*{ACKNOWLEDGMENT}
Parts of the code used to develop Fig.~\ref{fig:lintel_structure},~\ref{fig:lintel_build_sequence} and the results in this paper were developed by the authors in combination with Claude Sonnet and Opus models.


\bibliographystyle{IEEEtran}
\bibliography{base}

@article{Bruun21AIC_masonry,
  author  = {Bruun, E. P. G. and Pastrana, R. and Paris, V. and Beghini, A. and Pizzigoni, A. and Parascho, S. and Adriaenssens, S.},
  title   = {Three cooperative robotic fabrication methods for the scaffold-free construction of a masonry arch},
  journal = {Automation in Construction},
  volume  = {129},
  pages   = {103803},
  year    = {2021},
}

@article{Huang21CR_additive_bar,
  author  = {Huang, Y. and Garrett, C. R. and Ting, I. and Parascho, S. and Mueller, C. T.},
  title   = {Robotic additive construction of bar structures: Unified sequence and motion planning},
  journal = {Construction Robotics},
  volume  = {5},
  number  = {2},
  pages   = {115--130},
  year    = {2021},
}

@inproceedings{Peters25ICRA_assembly_voxels,
  author    = {Peters, T. and Cheung, K. and Kostitsyna, I.},
  title     = {Assembly order planning for modular structures by autonomous multi-robot systems},
  booktitle = {Proc. IEEE Int. Conf. Robotics and Automation (ICRA)},
  address   = {Atlanta, GA, USA},
  pages     = {11496--11502},
  year      = {2025},
  doi       = {10.1109/ICRA55743.2025.11127644},
}

@inproceedings{Culbertson19IROS_space_assembly,
  author    = {Culbertson, P. and Bandyopadhyay, S. and Schwager, M.},
  title     = {Multi-robot assembly sequencing via discrete optimization},
  booktitle = {Proc. IEEE/RSJ Int. Conf. Intelligent Robots and Systems (IROS)},
  address   = {Macau, China},
  pages     = {6502--6509},
  year      = {2019},
}

@article{Bruun22AIC_spaceframes,
  author  = {Bruun, E. P. G. and Adriaenssens, S. and Parascho, S.},
  title   = {Structural rigidity theory applied to the scaffold-free (dis)assembly of space frames using cooperative robotics},
  journal = {Automation in Construction},
  volume  = {141},
  pages   = {104405},
  year    = {2022},
  doi     = {10.1016/j.autcon.2022.104405},
}

@article{Chen22Arxiv_coop_task,
  author  = {Chen, J. and Li, J. and Huang, Y. and Garrett, C. and Sun, D. and Fan, C. and Hofmann, A. and Mueller, C. and Koenig, S. and Williams, B. C.},
  title   = {Cooperative task and motion planning for multi-arm assembly systems},
  journal = {arXiv preprint arXiv:2203.02475},
  year    = {2022},
}

@inproceedings{Hartmann20IROS_long_horz_plan,
  author    = {Hartmann, V. N. and Oguz, O. S. and Driess, D. and Toussaint, M. and Menges, A.},
  title     = {Robust task and motion planning for long-horizon architectural construction planning},
  booktitle = {Proc. IEEE/RSJ Int. Conf. Intelligent Robots and Systems (IROS)},
  pages     = {6886--6893},
  year      = {2020},
}

@inproceedings{Nagele20IROS_LegoBot,
  author    = {Nagele, L. and Hoffmann, A. and Schierl, A. and Reif, W.},
  title     = {{LegoBot}: Automated planning for coordinated multi-robot assembly of {LEGO} structures},
  booktitle = {Proc. IEEE/RSJ Int. Conf. Intelligent Robots and Systems (IROS)},
  pages     = {9088--9095},
  year      = {2020},
  doi       = {10.1109/IROS45743.2020.9341428},
}

@inproceedings{Cebula24IROS_feasible_assembly,
  author    = {Cebulla, A. and Asfour, T. and Kr{\"o}ger, T.},
  title     = {Beyond feasibility: Efficiently planning robotic assembly sequences that minimize assembly path lengths},
  booktitle = {Proc. IEEE/RSJ Int. Conf. Intelligent Robots and Systems (IROS)},
  address   = {Abu Dhabi, United Arab Emirates},
  pages     = {4076--4083},
  year      = {2024},
  doi       = {10.1109/IROS58592.2024.10801475},
}

@article{Kavraki96TRO_prm,
  author  = {Kavraki, L. E. and Svestka, P. and Latombe, J.-C. and Overmars, M. H.},
  title   = {Probabilistic roadmaps for path planning in high-dimensional configuration spaces},
  journal = {IEEE Transactions on Robotics and Automation},
  volume  = {12},
  number  = {4},
  pages   = {566--580},
  month   = {Aug.},
  year    = {1996},
}

@article{Yang26RCM_decent_plan,
  author  = {Yang, H. and Xu, W. and Pham, D. T. and Qi, L. and Ba, M.},
  title   = {Decentralised task planning and motion coordination for scalable multi-robot collaborative manufacturing},
  journal = {Robotics and Computer-Integrated Manufacturing},
  volume  = {100},
  pages   = {103255},
  year    = {2026},
  doi     = {10.1016/j.rcim.2026.103255},
}

@INPROCEEDINGS{Kuffner00ICRA_RRT_connect,
  author={Kuffner, J.J. and LaValle, S.M.},
  booktitle={Proceedings 2000 ICRA. Millennium Conference. IEEE International Conference on Robotics and Automation. Symposia Proceedings (Cat. No.00CH37065)}, 
  title={RRT-connect: An efficient approach to single-query path planning}, 
  year={2000},
  volume={2},
  number={},
  pages={995-1001 vol.2},
  doi={10.1109/ROBOT.2000.844730}}

@inproceedings{Gayle05RSS_deformable,
  author    = {Gayle, R. and Segars, P. and Lin, M. C. and Manocha, D.},
  title     = {Path planning for deformable robots in complex environments},
  booktitle = {Proc. Robotics: Science and Systems (RSS)},
  address   = {Cambridge, MA, USA},
  year      = {2005},
  doi       = {10.15607/RSS.2005.I.030},
}

@article{Aksoy26TASE_DLO,
  author  = {Aksoy, B. and Wen, J. T.},
  title   = {Planning and Control for Deformable Linear Object Manipulation},
  journal = {IEEE Transactions on Automation Science and Engineering},
  volume  = {23},
  pages   = {1093--1111},
  year    = {2026},
  doi     = {10.1109/TASE.2025.3635685},
}

@article{Yin21SciRob_deformable,
  author  = {Yin, H. and Varava, A. and Kragic, D.},
  title   = {Modeling, learning, perception, and control methods for deformable object manipulation},
  journal = {Science Robotics},
  volume  = {6},
  number  = {54},
  pages   = {eabd8803},
  year    = {2021},
  doi     = {10.1126/scirobotics.abd8803},
}

@techreport{Dellaert12GTSAM_factorgraphs,
  author      = {Dellaert, F.},
  title       = {Factor Graphs and {GTSAM}: A Hands-on Introduction},
  institution = {Georgia Institute of Technology},
  number      = {GT-RIM-CP\&R-2012-002},
  year        = {2012},
}

@article{Mukadam18IJRR_GPMP,
  author  = {Mukadam, M. and Dong, J. and Yan, X. and Dellaert, F. and Boots, B.},
  title   = {Continuous-time Gaussian process motion planning via probabilistic inference},
  journal = {The International Journal of Robotics Research},
  volume  = {37},
  number  = {11},
  pages   = {1319--1340},
  month   = {Sep.},
  year    = {2018},
  doi     = {10.1177/0278364918790369},
}

@book{Dellaert17_FactorGraphs,
  author  = {Frank Dellaert and Michael Kaess},
  title   = {{Factor Graphs for Robot Perception}},
  publisher = {Foundations and Trends in Robotics},
  journal = {Foundations and Trends in Robotics},
  volume  = {6},
  number  = {1--2},
  pages   = {1--139},
  year    = {2017},
  doi     = {10.1561/2300000043}
}

@inproceedings{Barfoot14RSS_BatchCT,
  title={Batch Continuous-Time Trajectory Estimation as Exactly Sparse Gaussian Process Regression},
  author={Tim D. Barfoot and Chi Hay Tong and Simo S{\"a}rkk{\"a}},
  booktitle={Robotics: Science and Systems Conference},
  year={2014},
  url={https://api.semanticscholar.org/CorpusID:16668561}
}

@article{Mukadam18CoRR_steap,
  author       = {Mustafa Mukadam and
                  Jing Dong and
                  Frank Dellaert and
                  Byron Boots},
  title        = {{STEAP:} simultaneous trajectory estimation and planning},
  journal      = {CoRR},
  volume       = {abs/1807.10425},
  year         = {2018},
  url          = {http://arxiv.org/abs/1807.10425},
  eprinttype   = {arXiv},
  eprint       = {1807.10425},
  bibsource    = {dblp computer science bibliography, https://dblp.org}
}

@INPROCEEDINGS{Alwala21IROS_jsmp,
  author={Alwala, Kalyan Vasudev and Mukadam, Mustafa},
  booktitle={2021 IEEE/RSJ International Conference on Intelligent Robots and Systems (IROS)}, 
  title={Joint Sampling and Trajectory Optimization over Graphs for Online Motion Planning}, 
  year={2021},
  volume={},
  number={},
  pages={4700-4707},
  doi={10.1109/IROS51168.2021.9636064}}

@article{Otte116IJRR_RRTX,
author = {Michael Otte and Emilio Frazzoli},
title ={RRTX: Asymptotically optimal single-query sampling-based motion planning with quick replanning},

journal = {The International Journal of Robotics Research},
volume = {35},
number = {7},
pages = {797-822},
year = {2016},
doi = {10.1177/0278364915594679},
URL = { 
    
        https://doi.org/10.1177/0278364915594679
},
eprint = { 
    
        https://doi.org/10.1177/0278364915594679
}
,
}

@INPROCEEDINGS{KingSmith22ICRA_SCATE,
  author={King–Smith, Matthew and Tsiotras, Panagiotis and Dellaert, Frank},
  booktitle={2022 International Conference on Robotics and Automation (ICRA)}, 
  title={Simultaneous Control and Trajectory Estimation for Collision Avoidance of Autonomous Robotic Spacecraft Systems}, 
  year={2022},
  volume={},
  number={},
  pages={257-264},
  doi={10.1109/ICRA46639.2022.9811875}}

@inproceedings{Thoma18ROBARCH_timber,
  author    = {Thoma, A. and Adel, A. and Helmreich, M. and Wehrle, T. and Gramazio, F. and Kohler, M.},
  title     = {Robotic fabrication of bespoke timber frame modules},
  booktitle = {Robotic Fabrication in Architecture, Art and Design 2018},
  pages     = {447--458},
  year      = {2018},
  publisher = {Springer International Publishing},
  doi       = {10.1007/978-3-319-92294-2_34},
}

@inproceedings{Parascho18AAG_spatial,
  author    = {Parascho, S. and Kohlhammer, T. and Coros, S. and Gramazio, F. and Kohler, M.},
  title     = {Computational design of robotically assembled spatial structures: A sequence based method for the generation and evaluation of structures fabricated with cooperating robots},
  booktitle = {Advances in Architectural Geometry 2018},
  pages     = {112--139},
  year      = {2018},
  publisher = {Klein Publishing GmbH},
}

@article{Bruun24CR_zerowaste,
  author  = {Bruun, E. P. G. and Besler, E. and Adriaenssens, S. and Parascho, S.},
  title   = {Scaffold-free cooperative robotic disassembly and reuse of a timber structure in the {ZeroWaste} project},
  journal = {Construction Robotics},
  volume  = {8},
  year    = {2024},
  doi     = {10.1007/s41693-024-00137-7},
}

@inproceedings{Macklin16ICMG_xpbd,
author = {Macklin, Miles and M{\"u}ller, Matthias and Chentanez, Nuttapong},
title = {XPBD: position-based simulation of compliant constrained dynamics},
year = {2016},
isbn = {9781450345927},
publisher = {Association for Computing Machinery},
address = {New York, NY, USA},
url = {https://doi.org/10.1145/2994258.2994272},
doi = {10.1145/2994258.2994272},
booktitle = {Proceedings of the 9th International Conference on Motion in Games},
pages = {49–54},
numpages = {6},
location = {Burlingame, California},
series = {MIG '16}
}

@misc{compas-fab,
    title={{COMPAS~FAB}: Robotic fabrication package for the COMPAS Framework},
    author={
        Rust, R. and
        Casas, G. and
        Parascho, S. and
        Jenny, D. and
        D\"{o}rfler, K. and
        Helmreich, M. and
        Gandia, A. and
        Ma, Z. and
        Ariza, I. and
        Pacher, M. and
        Lytle, B. and
        Huang, Y. and
        Kasirer, C. and
        Bruun, E. and
        Leung, P.Y.V.
        },
    howpublished={https://github.com/compas-dev/compas\_fab/},
    note={Gramazio Kohler Research, ETH Z\"{u}rich},
    year={2018},
    doi={10.5281/zenodo.3469478},
    url={https://doi.org/10.5281/zenodo.3469478},
}

@misc{compas-rrc,
    title  = {{COMPAS RRC}: Online control for ABB robots over a simple-to-use Python interface.},
    author = {Fleischmann, Philippe and Casas, Gonzalo and Lyrenmann, Michael},
    note   = {\url{https://compas-rrc.github.io/compas_rrc}},
    month  = {7},
    year   = {2020},
    doi    = {10.5281/zenodo.4639418},
    url    = {https://doi.org/10.5281/zenodo.4639418},
}

@article{Kaess12IJRR_isam2,
author = {Michael Kaess and Hordur Johannsson and Richard Roberts and Viorela Ila and John J Leonard and Frank Dellaert},
title ={iSAM2: Incremental smoothing and mapping using the Bayes tree},

journal = {The International Journal of Robotics Research},
volume = {31},
number = {2},
pages = {216-235},
year = {2012},
doi = {10.1177/0278364911430419},

URL = { 
    
        https://doi.org/10.1177/0278364911430419
    
    

},
eprint = { 
    
        https://doi.org/10.1177/0278364911430419
    
    

}
,
}

@misc{gtsam,
author       = {Frank Dellaert and {GTSAM Contributors}},
title        = {borglab/gtsam},
month        = may,
year         = {2022},
publisher    = {Georgia Tech Borg Lab},
note         = {version 4.2a8},
doi          = {10.5281/zenodo.5794541},
url          = {https://github.com/borglab/gtsam}
}

@article{Berrut04_Lagrange,
author = {Berrut, Jean-Paul and Trefethen, Lloyd N.},
title = {Barycentric Lagrange Interpolation},
journal = {SIAM Review},
volume = {46},
number = {3},
pages = {501-517},
year = {2004},
doi = {10.1137/S0036144502417715},

URL = { 
    
        https://doi.org/10.1137/S0036144502417715
    
    

},
eprint = { 
    
        https://doi.org/10.1137/S0036144502417715
    
    

}
,
}

\end{document}